\documentclass{llncs}

\usepackage{xspace}
\usepackage{latexsym}

\newcommand{\Thsld}{\mbox{\textsl{Th}$_{_{sld}}$}\xspace}

\newcommand{\Clarg}{\mbox{\textsl{C}$_{_{arg}}$}\xspace}
\newcommand{\Clwar}{\mbox{\textsl{C}$_{_{war}}$}\xspace}

\newcommand{\sld}{\mbox{\textsc{Sld}}\xspace}

\newcommand{\Pruned}[1]{$Pruned(#1)$}

\newcommand{\InferCred} 
 {\mbox{${\Infer}_{_{\vspace*{6mm}{\hspace*{-9pt}cred}}}$}}

\newcommand{\InferSL} 
  {\mbox{${\Infer}_{_{\vspace*{6mm}{\hspace*{-9pt}SL}}}$}}

\newcommand{\InferMTDR} 
  {\mbox{${\Infer}_{_{\vspace*{6mm}{\hspace*{-9pt}\textsl{MTDR}}}}$}}

\newcommand{\InferDelp} 
  {\mbox{${\Infer}_{_{\vspace*{6mm}{\hspace*{-9pt}\textsl{LP}}}}$}}

\newcommand{\InferNeg} 
  {\mbox{${\Infer}_{_{\vspace*{6mm}{\hspace*{-9pt}\textsl{neg}}}}$}}

\newcommand{\InferNot} 
  {\mbox{${\Infer}_{_{\vspace*{6mm}{\hspace*{-9pt}\textsl{not}}}}$}}

\newcommand{\InferNlp} 
  {\mbox{${\Infer}_{_{\vspace*{6mm}{\hspace*{-9pt}\textsl{NLP}}}}$}}

\newcommand{\SDEar}{\textsl{LDS}$_{AR}$\xspace}

\newcommand{\SDEskep}{\textsl{AS}$_{_{\mathsf{LP}}}$\xspace}

\newcommand{\SDEmtdr}{\textsl{AS}$_{_{\mathsf{MTDR}}}$\xspace}
  \newcommand{\SDEsl}{\textsl{AS}$_{_{\mathsf{SL}}}$\xspace}
 \newcommand{\SDEneg}{\textsl{AS}$_{_{\mathsf{neg}}}$\xspace}
 \newcommand{\SDEnot}{\textsl{AS}$_{_{\mathsf{not}}}$\xspace}
 \newcommand{\SDEnlp}{\textsl{AS}$_{_{\mathsf{NLP}}}$\xspace}  

\newcommand{\NLP}{\textsl{NLP}\xspace}

\newcommand{\DELP}{\textsl{DeLP}\xspace}

\newcommand{\DELPneg}{\ensuremath{\textsl{DeLP$_{neg}$}}\xspace}
\newcommand{\DELPnot}{\ensuremath{\textsl{DeLP$_{\textsf{not}}$}}\xspace}

\newcommand{\negacion}{$/\hspace*{-0.5em}$}

\newcommand{\comentario}[1]{}
\newcommand{\skippar}[1]{}

\newlength{\EA}  
\newlength{\MEA} 

\newlength{\premisas}
\newlength{\consec}
\newlength{\rotulo}
\newlength{\largoregla}

\newcommand{\ie}{\mbox{$i.e.$}\xspace}
\newcommand{\eg}{\mbox{$e.g.$}\xspace}

\newtheorem{Notax}{{\bf Observaci\'on}}[section]

\newtheorem{obsx}{{\bf Observaci\'on}}[section]

\newtheorem{Algox}{{\sc Algoritmo}}[section]

\newlength{\blk}
\newcommand{\srule}[2]{\ensuremath{\mathit{#1}\;\gets\;\mathit{#2}}\xspace}

\newcommand{\Srule}[2]{\ensuremath{(\mathit{#1}\gets\mathit{#2})}\xspace}

\newcommand{\Rimplies}{\mbox{$\leftarrow$}}

\newcommand{\Infer}{\mbox{$\mid\hspace{-2pt}\sim\ $}}

\newcommand{\Tree}[1]{\mbox{${\cal T}_{\scriptsize #1}$}}

\newcommand{\MTDR}{\mbox{\textsl{MTDR}}\xspace}

\newcommand{\SL}{\mbox{$\textsl{SL}$}\xspace}

\newcommand{\ArgumGeneral}[1]{\mbox{$\rangle {\cal A}, h \langle$}}

\newcommand{\mcup}{\mbox{$\cup$}}

\newcommand{\mGamma}{\mbox{$\Gamma$}}

\newcommand{\memptyset}{\mbox{$\emptyset$}}

\newcommand{\cto}[1]{\{#1\}}

\newcommand{\OrdenPref}{\mbox{$\preceq$}}

\newcommand{\Args}{{\sf Args}}

\newcommand{\Wffs}{{\sf Wffs}}

\newcommand{\LengKR}{\mbox{$\mathcal{L}_{_\mathsf{KR}}$}\xspace}

\newcommand{\LengLabel}{\mbox{$\mathcal{L}_{_\mathsf{Labels}}$}}

\newcommand{\LengArg}{\mbox{$\mathcal{L}_{_\mathsf{Arg}}$}\xspace}

\newcommand{\InferArg}
 {\mbox{${\Infer}_{_{\vspace*{6mm}{\hspace*{-9pt}Arg}}}$}}

\newcommand{\InferTree}
  {\mbox{${\Infer}_{_{\vspace*{6mm}{\hspace*{-9pt}{\small\cal T}}}}$}}

\newcommand{\TREE}[2]{\mbox{${\mathbf T}_{#1}^{#2}$}}
\newcommand{\UnTREE}{\mbox{${\mathbf T}$}}

\newcommand{\Argument}[1]{\mbox{${\cal A}_{#1}$}}

\newcommand{\du}[2]{\mbox{$#1$:$#2$}\xspace}

\newcounter{nro-regla}
\newenvironment{regla-infer}{\begin{center}}{\end{center}
                         \addtocounter{nro-regla}{1}}

\newcommand{\personal}[1]{}

\newcommand{\DeriProlog}{\mbox{$\vdash_{_{\tiny SLD}}\hspace*{0mm}$}\xspace}
\newcommand{\NotDeriProlog}{\mbox{$\negacion\DeriProlog$}}

\newcommand{\Argm}[2]{\du{{\cal #1}}{#2}}

\newcommand{\ArgSet}[1]{\mbox{${\cal #1}$}}

\newcommand{\ArgSeti}[2]{\mbox{${\cal #1}_{\scriptstyle #2}$}}

\newcommand{\ValidSubTree}{\mbox{\textsf{VSTree}}}

\newcommand{\Minimal}{\mbox{{\sf Minimal}}}

\newcommand{\no}{\ensuremath{\sim\!\!}\xspace}

\newcommand{\OpPi}{\mbox{$\mathbf\Pi$}}

\newcommand{\comentado}[1]{}

\newcommand{\IntroNR}{\mbox{{\sf Intro-NR}}}
\newcommand{\IntroRE}{\mbox{{\sf Intro-RE}}}
\newcommand{\IntroConj}{\mbox{{\sf Intro-$\wedge$}}}
\newcommand{\ElimImp}{\mbox{{\sf Elim-$\Rimplies$}}}

\newcommand{\IntroAD}{\mbox{{\sf Intro-1D}}}
\newcommand{\IntroNAD}{\mbox{{\sf Intro-ND}}}

\newcommand{\MarkAtomic}{\mbox{{\sf Mark-Atom}}}
\newcommand{\MarkD}{\mbox{{\sf Mark-1D}}}
\newcommand{\MarkND}{\mbox{{\sf Mark-ND}}}

\newcommand{\Pdft}{\mbox{$\succ$}}

\newcommand{\nulo}{*}

\newcommand{\antec}{\mbox{$\sqsubset$ }}

\newcommand{\via}{\mbox{via }}

\newcommand{\dur}[2]{ \du{ \etar{\cto{#1}}{\cto{#2}}}{ #1 } }
\newcommand{\dun}[2]{ \du{ \etar{\emptyset}{\cto{#2}}}{ #1 } }

\begin{document}

\begin{center}
     {\Large\bf      A Framework for Combining Defeasible \\ Argumentation
     with Labeled Deduction\protect\footnote{``A Framework for Combining Defeasible
     Argumentation with Labelled Deduction" (Carlos I. Ches\~nevar - Guillermo R. Simari).
     In ``Computer Modeling of Scientific Reasoning" (C.Delrieux, J.Legris, Eds.).
     Pp. 43-56, Ed. Ediuns, Argentina, 2003. ISBN 987-89281-89-6.}}
\end{center}


\begin{center}
{\bf Carlos Iv\'an Ches\~nevar}$^{^1}$     \hspace*{25mm} {\bf Guillermo Ricardo Simari}$^{^2}$ \\

\begin{footnotesize}
\begin{center}
\begin{tabular}{c}
$^1${Artificial Intelligence Research Group} --- {Departament of Computer Science}  \\
Universitat de Lleida -- Campus Cappont -- {C/Jaume II, 69 -- E-25001 Lleida, {\sc Spain}} \\
{\sc Tel/Fax:} (+34) (973) 70 2764 /  2702  -- {\sc Email:} {\tt cic@eup.udl.es} \\
\end{tabular}
\end{center}
\begin{center}
\begin{tabular}{c}
$^2${Artificial Intelligence Laboratory --- Dep. of Computer Science and Enginering}  \\
Universidad Nacional del Sur --  {Alem 1253 -- B8000CPB Bah\'{\i}a Blanca, {\sc Argentina}} \\
{\sc Tel/Fax:} (+54) (291) 459 5135/5136 -- {\sc Email:} {\tt grs@cs.uns.edu.ar} \\
\end{tabular}
\end{center}
\end{footnotesize}

\begin{footnotesize}
{\sc Key words:} {\sf Defeasible Argumentation, Defeasible Reasoning, Labelled Deduction}
\end{footnotesize}
\end{center}

\newcommand{\unrelated}{\mbox{$\sim$}}

\newcommand{\LengLabelPLUS}{\mbox{$\mathcal{L}_{_\mathsf{Labels}}^{*}$}}
\newcommand{\LengArgPLUS}{\mbox{$\mathcal{L}_{_\mathsf{Arg}}^{*}$}\xspace}

\newcommand{\SDEarPLUS}{\textsl{LDS}$_{AR}^{*}$\xspace}

\newcommand{\IntroN}{\mbox{\sf Intro-N}}
\newcommand{\IntroD}{\mbox{\sf Intro-D}}
\newcommand{\IntroAND}{\mbox{\sf Intro-$\wedge$}}
\newcommand{\MP}{\mbox{\sf MP}}

\nocite{ENGChesnePHD2001}
\nocite{Simari92}
\nocite{PraVre99}


\begin{abstract}
   In the last years, there has been an increasing demand of
   a variety of logical systems, prompted mostly by applications
   of logic in AI and other related areas.
   \emph{Labeled Deductive Systems} (LDS) were developed  as a flexible methodology
   to formalize such a kind of complex logical systems.

   \emph{Defeasible argumentation} has proven to be a successful approach
   to formalizing commonsense reasoning, encompassing
   many other alternative formalisms for defeasible reasoning.
   Argument-based frameworks share
   some common notions (such as the concept of argument, defeater,  etc.)
   along with a number of particular features which make it difficult to compare them
   with each other from a logical viewpoint.

   This paper introduces \SDEar, a LDS for defeasible argumentation
   in which many important issues concerning defeasible argumentation are
   captured within a unified logical framework.
   We also discuss some logical properties and extensions that
   emerge  from the proposed framework.
\end{abstract}

\section{Introduction and motivations}
\label{sec:intro}

Labeled Deductive Systems (LDS)~\cite{Gabbay96}  were developed as  a rigorous
but flexible methodology to formalize complex logical  systems, such as temporal logics,
database query languages and  defeasible reasoning systems.
In labeled deduction, the usual notion of  formula is replaced by the notion
of \emph{labeled formula}, expressed as   \emph{Label}:\emph{f},
where \emph{Label} represents a label associated with a  wff \emph{f}.
A labeling language \LengLabel\ and knowledge-representation
language \LengKR\ can be combined to provide a new, labeled language,
in which labels convey additional information also
encoded at object-language level.
Derived formulas are labeled according to a family of \emph{deduction rules},
and with agreed ways of propagating labels via the application of
these rules.

In the last decade \emph{defeasible argumentation}~\cite{Survey2000,PraVre99}
has proven to be a successful approach
   to formalizing commonsense reasoning, providing a suitable formalization that encompasses
   many other alternative formalisms. Thus, most argument-based frameworks share
   some common notions (such as the concept of argument, defeater, warrant, etc.)
   along with a number of particular features which make it difficult to compare them
   with each other from a logical viewpoint.

The study of logical properties of \emph{defeasible argumentation}
motivated the development of \SDEar~\cite{ENGChesnePHD2001,Tou2001,Cacic2001a},
an LDS-based argumentation formalism.
In \SDEar\ two languages \LengLabel\ (representing arguments and their interrelationships)
and \LengKR\ (representing object-level knowledge) are combined into
a single, labeled language \LengArg.
Inference rules are provided in \LengArg\ to characterize argument construction and
their relationships.
\SDEar provides thus a common framework for different purposes,
such as studying logical properties of defeasible argumentation,
comparing and analyzing  existing argument-based frameworks
and developing extensions of the original framework by enriching
the labeling language \LengLabel.

This paper is structured as follows.
First, in section~\ref{sec:intro_lds} we discuss the main definitions
and concepts associated with the \SDEar framework.
In Section~\ref{sec:logical} we present some logical properties
that hold in \SDEar, and show how different alternative
argument-based formalisms can be seen as particular instances of
the proposed framework.
Then in Section~\ref{sec:applications}  we discuss some particular
issues relating \SDEar to modeling scientific reasoning, such
as comparing top-down vs. bottom-up computation of warrant,
and the combination of qualitative and quantitative reasoning
by incorporating numerical attributes.
Finally Section~\ref{sec:conc} summarizes related work
as well as the main conclusions  that have been obtained.

\section{The \SDEar framework: fundamentals\protect\footnote{For
space reasons we only give a summary of the main elements of
the \SDEar framework; for an in-depth treatment see~\cite{ENGChesnePHD2001,Cacic2001a}).
We also assume that the reader has a basic knowledge about
defeasible argumentation formalisms~\cite{Survey2000,PraVre99}.}}

\label{sec:intro_lds}

\subsection{Knowledge representation}

We will first introduce a \emph{knowledge representation language}
\LengKR\  together with a \emph{labeling language} \LengLabel.
These languages will be used to define the object language \LengArg.
Following~\cite{Gabbay96}, labeled wffs in \LengArg\ will be called
\emph{declarative units}, having the form  \emph{Label:wff}.

\begin{definition}[Language \LengKR. Wffs in  \LengKR]
  The language  \LengKR\ will be composed of propositional atoms ($a$, $b$, \ldots)
  and the logical  connectives
  $\wedge$, $\no$\ \ and  $\Rimplies$.
  If $\alpha$ is an atom in  \LengKR, then  $\alpha$ and $\no$ $\alpha$ are wffs
  called  \emph{literals} in \LengKR.
  If $\alpha_1$, \ldots $\alpha_k$, $\beta$   are literals in $\LengKR$, then
  \srule{\beta}{\alpha_1, \ldots \alpha_k}  is a wff in  \LengKR\ called \emph{rule}.
\end{definition}

The language \LengKR\ is a Horn-like propositional language
restricted to  \emph{rules} and \emph{facts}.\footnote{The language \LengKR\ is similar
to the language of \emph{extended  logic programming} 
in a propositional setting.}
Labels in the language \LengLabel\ can be either \emph{basic} or
\emph{complex}.
Basic labels distinguish between defeasible and non-defeasible information,
whereas complex labels account for \emph{arguments}
(a tentative proof involving defeasible information)
and \emph{dialectical trees} (a tree-like structure rooted in a given argument).

\skippar{
A number of deduction rules characterizes a consequence
relation for performing inferences. Labels are propagated
via these deduction rules so that the set of all defeasible formulas
required in the proof  of a given literal $h$
are collected into a  \emph{set of support} $\mathcal{A}$.
Arguments are thus modeled as labeled
formulas $\mathcal{A}$:$h$, where
$\mathcal{A}$ stands for a set of ground (defeasible) clauses,
and $h$ (the \emph{conclusion} of the argument) is an extended literal.
As we will discuss later, this consequence relation
Arguments may be in conflict, resulting in a tree-like dialectical analysis
in which for a given  argument their defeaters, the defeaters of these
defeaters, and so on, are taken into account.
This will result in  new labeled wffs of the form
\emph{dialectical tree}:\emph{conclusion}.
Formally:
}

\begin{definition}[Labeling language  \LengLabel]
\label{def:label_language}
The labeling language  \LengLabel\ is a set of labels \{ $L_1$, $L_2$, \ldots $L_k$, \ldots\},
such that every label $L \in \LengLabel$ is:

\begin{enumerate}
\item The empty set $\emptyset$. This is a \emph{basic label} is associated with
 every wff which corresponds to non-defeasible knowledge.
\item A single wff $f$ in \LengKR. This is a \emph{basic label}
which corresponds to $f$ as a piece of defeasible knowledge.
\item A  set $\Phi \subseteq \Wffs(\LengKR)$. This is a complex label called \emph{argument label}.
\item A tree-like structure \TREE{}{} is a complex label called \emph{dialectical label}, being defined as follows:
      \begin{enumerate}
    \item  If $\Phi$ is an argument label,
       then
         $\TREE{}{U}(\Phi)$,
         $\TREE{}{D}(\Phi)$ and $\TREE{}{*}(\Phi)$
      are \emph{dialectical labels} in \LengLabel.
      For the sake of simplicity, we will write
      \TREE{k}{}\ to denote an arbitrary dialectical label.
     \item If \TREE{1}{}, \ldots, \TREE{k}{}\ are
      dialectical labels,
      then
      $\TREE{n}{U}(\TREE{1}{}, \ldots, \TREE{k}{})$,
      $\TREE{n}{*}(\TREE{1}{}, \ldots, \TREE{k}{})$
      and
      $\TREE{m}{D}(\TREE{1}{}, \ldots, \TREE{k}{})$
   will also be dialectical labels in \LengLabel.
\end{enumerate}
\item
    Nothing else is a label in \LengLabel.
\end{enumerate}
\end{definition}

The \emph{object (labeled) language} in \SDEar
is defined as  $\LengArg = (\LengLabel, \LengKR)$.
Since \LengKR is a Horn-like logic language, we will assume an underlying
inference mechanism $\DeriProlog$ equivalent
to \sld resolution~\cite{Lloyd87}, properly extended to handle
a negated literal $\no p$ as a new constant name
$no\_p$. Given $P \subseteq \Wffs(\LengKR)$,
we  write $P \DeriProlog \alpha$ to denote that $\alpha$
follows from $P$ via $\DeriProlog$.

\begin{definition}[Contradictory set of wffs in \LengKR]
\label{def:contradiction}
Given a set $S$ of wffs in \LengKR,
$S$ will be called a \emph{contradictory set}  (denoted $S\ \DeriProlog \bot$)
iff complementary literals $p$ and $\no p$
can be derived from $S$ via \DeriProlog.
\end{definition}

\emph{Basic} declarative units will be used to encode defeasible and
non-defeasible information available for an intelligent agent to reason
from a set $\Gamma$ of labeled wffs.
Such a set will be called \emph{argumentative theory}. Formally:

\begin{definition}[Basic declarative units. Argumentative theory]
\label{def:basic_dec_units}
A labeled wff \du{\psi}{\alpha}\ such that
$\alpha$ is a basic label (either (1) $\psi = \emptyset$ or (2) $\psi = \alpha$)
will be called a \emph{basic declarative unit} (bdu).
In case (1), the wff \du{\emptyset}{\alpha}\  will be called
a \emph{non-defeasible} bdu; in case (2),
the wff \du{\alpha}{\alpha}\  will be called
a \emph{defeasible} bdu.
A finite set
$\Gamma$ = \{ $\gamma_1$, $\gamma_2$, \ldots, $\gamma_k$\}
of bdu's will be called an \emph{argumentative theory}.
For every argumentative theory $\Gamma$ we will assume that the set of non-defeasible
formulas $\OpPi(\Gamma) = \{ \du{\emptyset}{\alpha} \mid \du{\emptyset}{\alpha} \in \Gamma \}$
is non-contradictory.
\end{definition}

Formulas with  an empty label correspond to `strict' knowledge.
Thus, \du{\emptyset}{p}\  and
\du{\emptyset}{\srule{p}{q}}\
stand for a fact $p$ and a logic programming clause \srule{p}{q}.
Defeasible facts (also known as \emph{presumptions})
and defeasible rules are represented by
formulas \du{\{p\}}{p}\  and  \du{\{\srule{p}{q}\}}{\srule{p}{q}}.
Thus, the classical default ``Birds typically fly"
will be represented in \SDEar as
\du{\{\srule{fly}{bird}\}}{\srule{fly}{bird}},
whereas the strict rule ``Penguins don't fly"
will be represented in \SDEar as
\du{\emptyset}{\srule{\no fly}{penguin}}.
Intuitively, the label of a bdu stands for an initial set of support associated
with a formula in the argumentative theory, and is used for consistency check
when performing inferences, as discussed later in Sec.~\ref{sec:argu}.

\begin{example}
\label{ejem:engine}
Consider an intelligent agent involved in controlling an engine with three
switches $sw1$, $sw2$ and $sw3$.
These switches regulate different features of the engine,
such as pumping system, speed, etc. Suppose
we have defeasible information about how this engine works.
\begin{itemize}
\item If the pump is clogged, then the engine gets no fuel.
\item When $sw1$  is on, normally fuel is pumped properly.
\item When fuel is pumped properly, fuel usually works ok.
\item When $sw2$ is on, usually oil is pumped.
\item When oil is pumped, usually it works ok.
\item When there is oil and fuel, usually the engine works ok.
\item When there is fuel, oil, and heat, then the engine
      is usually not ok.
\item When there is heat, normally there are oil problems.
\item When fuel is pumped and speed is low,
      then there are reasons to believe that the pump is clogged.
\item When $sw2$ is on, usually speed is low.
\item When $sw3$ is on, usually fuel is ok.
\end{itemize}
Suppose we also know  some particular facts:
$sw1$, $sw2$ and $sw3$ are on, and there is heat.
The knowledge of such an agent can be modeled by the argumentative
theory $\Gamma_{engine}$ shown in figure~\ref{fig:ejem_maquina}.
$\Box$
\end{example}


\renewcommand{\dur}[1]{ \du{\{#1\}}{#1} }


\renewcommand{\dun}[1]{ \du{\emptyset}{#1} }

\newcommand{\espacio}{}

\begin{figure}[t]
\begin{center}
\begin{footnotesize}
         \begin{tabular}[b]{|l|} \hline
       \ \\
           \espacio \dun{ \srule{\no fuel\_ok}{pump\_clog}  }\\
          \espacio \dun{ \srule{sw1}{}                        }\\
           \espacio             \dun{ \srule{sw2}{}                        } \\
           \espacio             \dun{ \srule{sw3}{}                        } \\
           \espacio             \dun{ \srule{heat}{}                       } \\
           \espacio             \dur{\srule{pump\_fuel\_ok}{sw1}           } \\
           \espacio             \dur{\srule{fuel\_ok}{pump\_fuel\_ok }     } \\
           \espacio             \dur{\srule{pump\_oil\_ok}{sw2}            } \\
           \espacio             \dur{\srule{oil\_ok}{pump\_oil\_ok}        } \\
           \espacio             \dur{\srule{engine\_ok}{fuel\_ok,oil\_ok } } \\
           \espacio             \dur{\srule{\no engine\_ok}{fuel\_ok,oil\_ok,heat}     } \\
           \espacio             \dur{\srule{\no oil\_ok}{heat}                         }\\
           \espacio             \dur{\srule{pump\_clog}{pump\_fuel\_ok, low\_speed}  } \\
           \espacio             \dur{\srule{low\_speed}{sw2}                            } \\
           \espacio             \dur{\srule{\no low\_speed}{sw2,sw3}                   } \\
           \espacio             \dur{\srule{fuel\_ok}{sw3}                              } \\
           \espacio        \ \\ \hline
        \end{tabular}
\end{footnotesize}
\end{center}
\caption{Argumentative theory $\Gamma_{engine}$ (example~\protect\ref{ejem:engine})}
\label{fig:ejem_maquina}
\end{figure}


\subsection{Argument construction}
\label{sec:argu}

Given an argumentative theory $\Gamma$, and a wff $p \in \LengKR$,
the inference process in \SDEar\ involves first obtaining
a tentative proof (or \emph{argument}) for $p$.
A consequence relation $\InferArg$ propagates labels, implementing the
SLD resolution procedure along with a consistency check every time
new defeasible information is introduced in a proof.
Figure~\ref{fig:rules}
summarizes the natural deduction rules which characterize
the inference relationship $\InferArg$.
Rules \IntroNR\ and \IntroRE\  allow the introduction of
non-defeasible and defeasible information in a proof, respectively.
Rules \IntroConj\ and \ElimImp\ stand for introducing conjunction
and applying modus ponens.
In the last three rules, a consistency check is performed in order
to ensure that the label $\mathcal{A}$
together with \OpPi(\mGamma)
does not derive complementary literals, avoiding logical contradiction.
Note that the label $\mathcal{A}$ associated with a formula $\mathcal{A}$:$h$
contains all \emph{defeasible} information needed to conclude $h$ from $\Gamma$.
Thus, arguments in \SDEar\ are modeled as labeled
formulas $\mathcal{A}$:$h$, where
$\mathcal{A}$ stands for a set of (ground) defeasible rules
that along with \OpPi(\mGamma)  derive $h$.

\setlength{\unitlength}{1mm}
\begin{figure}[t]

\begin{picture}(120,40)
 \put(0,0){\framebox(120,40){
  \begin{minipage}{120mm}
    \begin{enumerate}
    \item  \IntroNR: \ \ \
    \begin{tabular}{c}
       \\  \hline
            \du{\memptyset}{\alpha}
    \end{tabular}
    for any $\du{\memptyset}{\alpha}$ 

    \item \IntroRE: \ \ \
    \begin{tabular}{c}
       \OpPi(\mGamma) \mcup\ $\Phi$ \NotDeriProlog $\bot$ \\ \hline
       \du{\Phi}{\alpha}
    \end{tabular}
    \noindent
    for any $\du{\Phi}{\alpha}$ 

    \item \IntroConj: \ \ \
    \begin{tabular}{c}
    \du{ \Phi_1 } {\alpha_1}  \ \ \ \
    \du{ \Phi_2 } {\alpha_2} \ \ \
    \ldots \du{ \Phi_k } {\alpha_k} \ \ \
    $\OpPi(\Gamma)$ $\cup$\ $\bigcup_{i=1\ldots k} \Phi_i \NotDeriProlog \bot$ \\ \hline
    \du{ \bigcup_{i=1\ldots k} \Phi_i }
       {\alpha_1, \alpha_2, \ldots, \alpha_k}
    \end{tabular}

    \item \ElimImp: \ \ \
    \begin{tabular}{c}
    \du{ \Phi_1}{\beta \Rimplies \alpha_1,\ldots,\alpha_k}
    \ \ \
    \du{ \Phi_2 }{\alpha_1,\ldots,\alpha_k}
    \ \ \ \
    $\OpPi(\Gamma)$ $\cup$\ $\Phi_1 \cup\ \Phi_2 \NotDeriProlog \bot$ \\ \hline
    \du{ \Phi_1 \cup \Phi_2}{\beta}
    \end{tabular}
    \end{enumerate}

  \end{minipage}
}}
\end{picture}

\caption{Inference rules for \InferArg: deriving (generalized) arguments in \SDEar}
\label{fig:rules}
\end{figure}

\begin{definition}[Argument. Subargument]
\label{DefGeneralizedArgument}  \label{def:concarg}
\label{DefArgumentoGeneralizado}
Let $\Gamma$ be an argumentative theory, and let
$h$ be a literal such that $\Gamma \InferArg \du {\ArgSet{A}} {h}$
Then \ArgSet{A}\ will be called a \emph{generalized argument} for $h$.
If it is not the case that $\Gamma \InferArg \du { \ArgSet{B} }{h}$, with $\ArgSet{B} \subset \ArgSet{A}$,
then \du {\ArgSet{A}} {h}\ is called a \emph{minimal argument} or just \emph{argument}.
If $\Gamma \InferArg \du {\ArgSet{A}} {h}$, and \du {\ArgSet{A}} {h} is an argument,
we will also say that \du {\ArgSet{A}} {h} is an argument \emph{based on} $\Gamma$

\noindent
An argument
\du {\ArgSet{A}} {h}\
is a \emph{subargument} of another argument
\du {\ArgSet{B}} {q}\
if $\ArgSet{A} \subset \ArgSet{B}$.

\end{definition}


\begin{example}
\label{example:arguments}
Consider the argumentative theory $\Gamma_{engine}$ from example~\ref{ejem:engine}.
Then the argument \du{\ArgSet{A}}{engine\_ok}, with
\begin{center}
\begin{tabular}{lcp{10cm}}
\ArgSet{A} & = & \{ \Srule{pump\_fuel\_ok}{sw1},
                    \Srule{pump\_oil\_ok}{sw2},
                    \Srule{fuel\_ok}{pump\_fuel\_ok},
                    \Srule{oil\_ok}{pump\_oil\_ok},
                    \Srule{engine\_ok}{fuel\_ok, oil\_ok} \} \\
\end{tabular}
\end{center}

\noindent
can be inferred via \InferArg\ by applying
the inference rules \IntroNR\ twice (inferring  $sw1$ and $sw2$),
then \IntroRE\ twice (inferring
\srule{pump\_fuel\_ok}{sw1} and
                \srule{pump\_oil\_ok}{sw2}),
then \IntroRE\ twice again to infer
                    \srule{fuel\_ok}{pump\_fuel\_ok} and
                    \srule{oil\_ok}{pump\_oil\_ok},
and finally
\IntroRE\ once again to infer
\srule{engine\_ok}{fuel\_ok, oil\_ok}.
In a similar way, arguments \du{\ArgSet{B}}{\no fuel\_ok},
\du{\ArgSet{C}}{\no low\_speed},
\du{\ArgSet{D}}{fuel\_ok}\ and
\du{\ArgSet{E}}{\no engine\_ok} can be derived via \InferArg, with
\begin{center}

\begin{tabular}{lcp{10cm}}
\ArgSet{B} & = &  \{
                    \Srule{pump\_fuel\_ok}{sw1},
                    \Srule{low\_speed}{sw2},
                    \Srule{pump\_clog}{pump\_fuel\_ok, low\_speed}  \}  \\
\end{tabular}

\begin{tabular}{lcp{10cm}}
\ArgSet{C} & = &   \{ \Srule{{\no}\ low\_speed}{sw2, sw3} \} \\
\end{tabular}

\begin{tabular}{lcp{10cm}}
\ArgSet{D} & = &   \{ \Srule{{\no}\ low\_speed}{sw2, sw3} \} \\
\end{tabular}

\begin{tabular}{lcp{10cm}}
\ArgSet{E} & = &   \{    \Srule{pump\_fuel\_ok}{sw1},
                  \Srule{pump\_oil\_ok}{sw2},
                  \Srule{fuel\_ok}{pump\_fuel\_ok},
                  \Srule{oil\_ok}{pump\_oil\_ok},
                  \Srule{\no engine\_ok}{fuel\_ok, oil\_ok, heat}\} \\
\end{tabular}
\end{center}
$\Box$
\end{example}

\begin{figure}[t]
\begin{picture}(120,80)
 \put(0,0){\framebox(120,80){
  \begin{minipage}{120mm}
\begin{enumerate}
\item \IntroAD:   \vspace*{-.3cm}
          \begin{regla-infer}
            \begin{tabular}{c}
              \du { \ArgSet{A}} {h}   \ \ \ $\Minimal(\du { \ArgSet{A}} {h} )$ \\ \hline
               \du{\TREE{}{*}(\ArgSet{A})}{h}
            \end{tabular}
          \end{regla-infer}

\item \IntroNAD: \vspace*{-.1cm}
      \begin{regla-infer}
         \begin{tabular}{c}
           \du{\TREE{}{*}(\ArgSet{A})}{h} \ \ \
                 \du{\TREE{1}{*}(\ArgSeti{B}{1},  \ldots)}{q_1} \ \
                 \du{\TREE{k}{*}(\ArgSeti{B}{k},  \ldots)}{q_k} \ \ \ \ \
                 \ValidSubTree(\ArgSet{A}, \TREE{i}{*}), $i=1 \ldots k$  \\  \hline
           \du{\TREE{}{*}(\ArgSet{A}, \TREE{1}{*}, \ldots, \TREE{k}{*}) }{h}
         \end{tabular}
      \end{regla-infer}

\item \MarkAtomic:
\vspace*{-.3cm}
\begin{regla-infer}
  \begin{tabular}{c}
     \du{\TREE{}{*}(\ArgSet{A})}{h} \\ \hline
     \du{[\TREE{}{U}(\ArgSet{A})]}{h}
  \end{tabular}
\end{regla-infer}

\item \MarkD:
\noindent
for some \TREE{i}{*}, $i=1\ldots k$
\begin{regla-infer}
\begin{tabular}{c}
\du{ [\TREE{}{*}( \ArgSet{A}, \TREE{1}{*}, \ldots,\TREE{i}{*},\ldots,
                                 \TREE{k}{} )] }{h}
\ \ \
\du{ [\TREE{i}{U}(\ArgSeti{B}{i}\ldots )]}{q_i} \ \ \ \ \  \ValidSubTree(\ArgSet{A},\TREE{i}{U}) \\ \hline
\du{[ \TREE{}{D}( \ArgSet{A}, \TREE{1}{*}, \ldots,\TREE{i-1}{*}, \TREE{i}{U}, \TREE{i+1}{*}, \ldots, \TREE{k}{*} )] }{h} \\
\end{tabular}
\end{regla-infer}

\item \MarkND:
\noindent
For all \TREE{i}{*}, $i=1\ldots k$
\begin{regla-infer}
\begin{tabular}{c}
\du{ [\TREE{}{*}( \Argument{}, \TREE{1}{*}, \ldots, \TREE{i}{*},\ldots, \TREE{k}{*} )] }{h}
\ \ \ \du{[\TREE{i}{D}(\ArgSeti{B}{i},\ldots)]}{q_i} \ \ \ \ \  \ValidSubTree(\ArgSet{A},\TREE{i}{D}) \\ \hline
\du{ [\TREE{}{U}( \Argument{}, \TREE{1}{D}, \ldots,\TREE{i}{D}, \ldots, \TREE{k}{D})] }{h} \\
\end{tabular}
\end{regla-infer}
\end{enumerate}

  \end{minipage}
}}
\end{picture}

\caption{Rules for building dialectical trees in \SDEarPLUS}
\label{fig:rules_trees}
\end{figure}

\subsection{Defeat among Arguments. Warrant}
\label{sec:defeat_and_war}

Given an argument $\du{\ArgSet{A}}{h}$ based on an argumentative theory $\Gamma$,
there may exist other conflicting arguments based on $\Gamma$ that
\emph{defeat} it.
Conflict among arguments is captured by the notion of contradiction
(def.~\ref{def:contradiction}).

\begin{definition}[Counterargument]
Let $\Gamma$ be an argumentative theory, and let $\du{\ArgSet{A}}{h}$
and $\du{\ArgSet{B}}{q}$ be arguments based on $\Gamma$.
Then \du{\ArgSet{A}}{h}\ \emph{counter-argues} \du{\ArgSet{B}}{q}\ if
there exists a subargument \du{\ArgSet{B'}}{s}\ of  \du{\ArgSet{B}}{q}\
such that
$\OpPi(\mGamma) \cup \cto{h,s}$ is contradictory.
The argument \du{\ArgSet{B'}}{s}\ will be called
\emph{disagreement subargument}.
\end{definition}

Defeat among arguments involves a  partial order which establishes a
\emph{preference criterion}  on conflicting arguments.
A common preference criterion is specificity~\cite{Simari92,SpecArticle2003},
which favors an argument with greater information content and/or less
use of defeasible rules.

\begin{definition}[Preference order \OrdenPref]
\label{DefOrdenPreferencia}
Let $\Gamma$ be an argumentative theory, and
let $\Args(\Gamma)$ be the set of arguments that can be obtained from
$\Gamma$.
A \emph{preference order} \OrdenPref\ $\subseteq$
$\Args(\Gamma) \times \Args(\Gamma)$ is any partial order on $\Args(\Gamma)$.
\end{definition}

\begin{definition}[Defeat]
\label{DefDerrota}  \label{def:derrota}
Let $\Gamma$ be an argumentative theory, such that
$\Gamma \InferArg \du{\ArgSet{A}}{h}$
and
$\Gamma \InferArg \du{\ArgSet{B}}{q}$.
We will say that
\du{\ArgSet{A}}{h}\ \emph{defeats} \du{\ArgSet{B}}{q}\
(or equivalently \du{\ArgSet{A}}{h}\ is a \emph{defeater}
for \du{\ArgSet{B}}{q}) if
\begin{enumerate}
\item
   \du{\ArgSet{A}}{h}\ counterargues  \du{\ArgSet{B}}{q},
   with disagreement subargument \du{\ArgSet{B'}}{q'}.
\item
       Either it holds that \du{\ArgSet{A}}{h} \Pdft\ \du{\ArgSet{B'}}{q'}, or
       \du{\ArgSet{A}}{h}\  and \du{\ArgSet{B'}}{q'}\ are unrelated
      by the preference order ``\OrdenPref".
\end{enumerate}
\end{definition}

\begin{example}
\label{example:defeaters}
Consider the argumentative theory from example~\ref{ejem:engine}.
Note that the arguments \du{\ArgSet{B}}{\no fuel\_ok},
and \du{\ArgSet{E}}{\no engine\_ok},
are counter-arguments for the original argument
\du{\ArgSet{A}}{engine\_ok},
whereas
\du{\ArgSet{C}}{\no low\_speed}\ and
\du{\ArgSet{D}}{fuel\_ok}\ are counter-arguments for
\du{\ArgSet{B}}{\no fuel\_ok}.
In each of these cases, counter-arguments are also
defeaters according to the specificity preference criterion~\cite{Simari92}.
\end{example}

Since defeaters are arguments, there may exist defeaters for the
defeaters and so on. That prompts for a complete dialectical analysis
to determine which arguments are ultimately defeated.

\begin{definition}[Dialectical Tree]
\label{def:dial_tree}
  Let $\mathcal{A}$ be an argument for $q$. The dialectical tree for
  \du{\mathcal{A}}{q}, denoted \Tree{\du{\mathcal{A}}{q}}, is
  recursively defined as follows:
\begin{enumerate}
\item A single node labeled with an argument \du{\mathcal{A}}{q} with
  no defeaters is by itself the dialectical tree
  for \du{\mathcal{A}}{q}.
\item Let $\du{\mathcal{A}_1}{q_1}, \du{\mathcal{A}_2}{q_2}, \ldots,
  \du{\mathcal{A}_n}{q_n}$ be all the defeaters 
  for \du{\mathcal{A}}{q}. We construct the dialectical tree for
  \du{\mathcal{A}}{q}, \Tree{\du{\mathcal{A}}{q}}, by labeling the
  root node with \du{\mathcal{A}}{q} and by making this node the
  parent node of the roots of the dialectical trees for $\du{{\cal
      A}_1}{q_1}, \du{\mathcal{A}_2}{q_2}, \ldots, \du{{\cal
      A}_n}{q_n}$.
\end{enumerate}
\end{definition}

\noindent
\textbf{Note}: in order to avoid {\em fallacious
argumentation\/}~\cite{Chile94}, some additional constraints not given
in Def.~\ref{def:dial_tree} are imposed on every path (e.g. there can be no
repeated arguments, as this would lead to circular argumentation).\footnote{An in-depth analysis is outside
the scope of this paper. See~\cite{Survey2000,Chile94} for  details.}

A dialectical tree resembles a \emph{dialogue tree} between two parties, proponent and opponent.
Branches of the tree correspond to exchange of arguments between these two parties.
A dialectical tree can be marked as an \textsc{and-or} tree~\cite{Ginsberg93}
according to the following procedure:
nodes with no defeaters (leaves) are marked as $U$-nodes (undefeated nodes).
Inner nodes are marked as $D$-nodes (defeated  nodes) iff they have at least
one $U$-node as a child, and as $U$-nodes iff they have every
child marked as $D$-node. Formally:

\begin{definition}[Marking of the Dialectical Tree]
  Let \du{\mathcal{A}}{q} be an argument and \Tree{\du{{\cal
        A}}{q}} its dialectical tree, then:
\begin{enumerate}
\item All the leaves in \Tree{\du{\mathcal{A}}{q}} are labeled as
  $U$-nodes.
\item Let \du{{\cal B}}{h} be an inner node of \Tree{\du{{\cal
        A}}{q}}. Then \du{{\cal B}}{h} will be a $U$-node iff every
  child of \du{{\cal B}}{h} is a $D$-node. The node \du{{\cal
      B}}{h} will be a $D$-node iff it has at least one child marked as
  $U$-node.
\end{enumerate}
\end{definition}

After performing the above dialectical analysis, an argument $\mathcal{A}$ which turns to
be ultimately undefeated is called a {\em warrant}.  Formally:

\begin{definition}[Warrant]
  Let \du{\mathcal{A}}{q} be an argument and \Tree{\du{{\cal
        A}}{q}} its associated dialectical tree, such that its root node
\du{{\cal    A}}{q} is marked as $U$. Then
\du{\mathcal{A}}{q} is called
a {\em warranted argument} or just {\em warrant}
\end{definition}

In the context of \SDEar, the construction and marking
of dialectical trees is captured in terms of \emph{dialectical labels} (Def.~\ref{def:label_language}).
Special marks (\nulo, $U$, $D$) are associated
with the a label $\UnTREE(\ArgSet{A}, \ldots)$  in order to determine whether
$\ArgSet{A}$ correspond to an
\emph{unmarked}, \emph{defeated} or \emph{undefeated} argument,
resp.
In the theory of defeasible argumentation,
a warranted argument or belief will be that one
which is ultimately accepted at some time of the dialectical process.
In \SDEar\ the concept of warrant can be formalized as follows:

\begin{definition}[Warrant -- Version 1]
\label{def:warrant_preliminar}
Let $Cn_{*}^{k}(\Gamma)$ be the set of all dialectical formulas
that can be obtained from $\Gamma$ via \InferTree\ by at most $k$ applications
of inference rules ($i <=k$).
A literal $h$ is said to be \emph{warranted} iff
$\du{\TREE{}{U}(\ArgSet{A},...)}{h} \in Cn_{*}^{k}(\Gamma)$,
and there is no $k'>k$, such that
$\du{\TREE{}{D}(\ArgSet{A},...)}{h} \in ( Cn_{*}^{k'}(\Gamma) \setminus Cn_{*}^{k}(\Gamma))$.
\end{definition}

This approach resembles Pollock's original ideas of (ultimately)
justified belief~\cite{Pollock95}. Note that Def.~\ref{def:warrant_preliminar}
forces to compute the closure under \InferTree in order to determine whether
a literal is warranted or not. Fortunately this is not the case, since
warrant can be captured in terms of a \emph{precedence relation} ``\antec''
between dialectical labels.
Informally, we will write \UnTREE\  \antec  \UnTREE'
whenever \UnTREE\ reflects a state in a dialogue which is
previous to \UnTREE' (in other words, \UnTREE' stands for
a dialogue which evolves from \UnTREE\ by incorporating new
arguments). A \emph{final label} is a label that cannot be further
extended.

\begin{definition}[Warrant -- Version 2] \footnote{It can be proven that Def.~\ref{DefWarrant}
and~\ref{def:warrant_preliminar}  are equivalent~\cite{ENGChesnePHD2001}.}
\label{DefWarrant}
Let $\Gamma$ be an argumentative theory, such that
$\Gamma$ \InferTree \du{\TREE{i}{U}(\ArgSet{A}, \ldots)}{h}
and \TREE{i}{U}\ is a final label
(\ie, it is not the case that
 $\Gamma$ \InferTree \du{\TREE{j}{D}(\ArgSet{A}, \ldots)}{h} \ \
and  \TREE{i}{U} \antec \TREE{j}{D}).
Then
\du{\TREE{i}{U}(\ArgSet{A}, \ldots)}{h}
is a \emph{warrant}. We will also say that
$h$ is a warranted literal, or 
that \du{\ArgSet{A}}{h}\ is a \emph{warrant} in $\Gamma$.
\end{definition}



In \SDEar, the construction of dialectical trees is formalized in terms of an
inference relationship \InferTree.
Figure~\ref{fig:rules_trees} summarizes the
rules needed for formalizing the above dialectical analysis.
Rule \IntroAD\ allows to generate a tree with a single argument
(\ie, a generalized argument which is minimal).
Rule \IntroNAD\ allows to expand a given tree \TREE{}{*}\
by introducing new subtrees
\du{\TREE{1}{*}(\ArgSeti{B}{1},  \ldots)}{q_1} \ \
                 \du{\TREE{k}{*}(\ArgSeti{B}{k},  \ldots)}{q_k}.
A special condition
\ValidSubTree(\ArgSet{A}, \TREE{i}{*}) , $i = 1 \ldots k$
checks that such subtrees are valid (\ie the root of every \TREE{i}{*}\ is
a defeater for the root of \TREE{}{*}, and
no fallacious argumentation is present).
Rules \MarkAtomic, \MarkD\ and \MarkND\ allow to `mark' the nodes (arguments)
in a dialectical tree as defeated  or undefeated.
The tree is marked as an \textsc{and-or} tree.
Nodes with no defeaters are marked as $U$-nodes (undefeated nodes).
Inner nodes are marked as $D$-nodes (defeated  nodes) iff they have at least
one $U$-node as a child, and as $U$-nodes iff they have every
child marked as $D$-node.


\begin{example}
\label{example:dial}
Consider the argumentative theory from example~\ref{ejem:engine}
and the arguments and defeat relations from examples~\ref{example:arguments}
and~\ref{example:defeaters}.
From the argumentative theory $\Gamma_{engine}$
the following formulas can be inferred via \InferTree:

\[
\begin{array}{lll}
\Gamma \InferTree  \du{\TREE{1}{*}(\ArgSet{A})}{engine\_ok}             &  \via \IntroAD & (1) \\
\Gamma \InferTree  \du{\TREE{2}{*}(\ArgSet{B})}{\no fuel\_ok}           &  \via \IntroAD & (2) \\
\Gamma \InferTree  \du{\TREE{3}{*}(\ArgSet{C})}{\no low\_speed}         &  \via \IntroAD & (3) \\
\Gamma \InferTree  \du{\TREE{4}{*}(\ArgSet{D})}{fuel\_ok}               &  \via \IntroAD & (4) \\
\Gamma \InferTree  \du{\TREE{5}{*}(\ArgSet{E})}{\no engine\_ok}         &  \via \IntroAD & (5) \\
\Gamma \InferTree  \du{\TREE{2}{*}(\ArgSet{B},
                                       \TREE{3}{*}(\ArgSet{C}),
                                       \TREE{4}{*}(\ArgSet{D})
                                    )}{\no fuel\_ok} &  \via \IntroNAD, (3) \textrm{ and } (4)  &  (6) \\
\Gamma \InferTree  \du{\TREE{1}{*}(\ArgSet{A},
                                     \TREE{2}{*}(\ArgSet{B},
                                                    \TREE{3}{*}(\ArgSet{C}),
                                                    \TREE{4}{*}(\ArgSet{D})
                                                ),
                                     \TREE{5}{*}(\ArgSet{E}
                                                )
                                     )}{engine\_ok} &  \via \IntroNAD \textrm{ and } (6)    &  (7)  \\
\Gamma \InferTree  \du{\TREE{5}{U}(\ArgSet{E})}{\no engine\_ok}         &  \via \MarkAtomic & (8) \\
\Gamma \InferTree  \du{\TREE{1}{D}(\ArgSet{A},
                                     \TREE{2}{*}(\ArgSet{B},
                                                    \TREE{3}{*}(\ArgSet{C}),
                                                    \TREE{4}{*}(\ArgSet{D})
                                                ),
                                     \TREE{5}{U}(\ArgSet{E}
                                                )
                                     )}{engine\_ok} &  \via \MarkD \textrm{ and } (8)    &  (9)  \\
\end{array}
\]

Note that the formula obtained in step (7) has a final label associated with it,
since it cannot be `expanded' from previous formulas. Hence,
following definition~\ref{DefWarrant}, we can conclude that
$engine\_ok$
is not warranted.
\end{example}

\section{\SDEar: Some relevant logical properties}

\label{sec:logical}



\SDEar\ provides a useful formal framework for
studying logical properties of argument-based systems
in terms of \emph{inference relationships}.\footnote{See~\cite{Antoniou} for an excellent survey
on the role of inference relationships and their properties in nonmonotonic logics.}
Three particular consequence operators can be identified:
\begin{itemize}
\item $Th_{sld}(\Gamma)$,
= $\{ \du{\emptyset}{h} $ $\mid$ $\Gamma \InferArg \du{\emptyset}{h}  \}$,
which denotes the set of non-defeasible conclusions that
follow from $\Gamma$ by using only strict rules.
\item
$\Clarg$ =  $ \{ \du{\ArgSet{A}}{\alpha}$ $\mid$  $\Gamma \InferArg \du{\ArgSet{A}}{\alpha}$,
where
$\alpha$ is a literal in \LengKR \}, which denotes the set of all arguments that follow from $\Gamma$;
\item
$\Clwar$ =  $ \{ \du{\emptyset}{h}$ $\mid$   {\rm there exists a warranted argument}
 \du{\ArgSet{A}}{h}\ {\rm based on} $\Gamma$ \},
which denotes the set of all warranted conclusions that follow from $\Gamma$;
\end{itemize}


Cummulativity was proven to hold for argumentative formulae. This
allows to think of an  argumentative theory 
as a knowledge base containing
`atomic' arguments (facts and rules), which can be later on
extended by incorporating new, more complex arguments.
Cummulativity is proven \emph{not} to hold for warranted conclusions,
following the intuitions suggested by Prakken \& Vreeswijk~\cite{PraVre99}.

\begin{lemma}[Cummulativity for Arguments]\footnote{Proofs of propositions and theorems
are not included for space reasons.
For details the interested reader is referred to~\cite{ENGChesnePHD2001,Cacic2001a,Cacic2002Combining}.}
\label{LemaCumulatividadArgum}
\label{lema:cumu}
Let $\Gamma$ be an argumentative theory, and let
$\alpha_1$ and $\alpha_2$ be wffs in \LengKR.
Then
$\Gamma \InferArg \Argm{\ArgSeti{A}{1}}{\alpha_1}$
implies that
  $\Gamma\ \cup\ \{ \Argm{\ArgSeti{A}{1}}{\alpha_1} \}$ \InferArg
\Argm{\ArgSeti{A}{2}}{\alpha_2}\ \  iff \ \
$\Gamma\ \InferArg \Argm{\ArgSeti{A}{2}}{\alpha_2}$
\end{lemma}

A special variant of \emph{superclassicality}
was shown to hold for both argument construction
and warrant wrt SLD resolution: if $Th_{sld}(\Gamma)$
denotes the set of conclusions that can be obtained from $\Gamma$
via  SLD, then it holds that
$\Clarg(\Gamma) \subseteq Th_{sld}(\Gamma)$
and
$\Clwar(\Gamma) \subseteq Th_{sld}(\Gamma)$,
where
$\Clarg$ and $\Clwar$  stand for the consequence operator
for argument construction and warrant, respectively.
This implies, among other things, that the analysis of attack between
arguments can be  focused on literals in defeasible rules.
Formally:

\begin{lemma}[Horn supraclassicality for $\Clarg$ and $\Clwar$]
\label{def:superclasi_horn_inferarg}
Operators $\Clarg(\Gamma)$
and $\Clwar$
satisfy Horn supraclassicality wrt
\Thsld, \ie  $\Thsld(\Gamma) \subseteq \Clarg(\Gamma)$
and $\Thsld(\Gamma) \subseteq \Clwar(\Gamma)$.
\end{lemma}

Analogously, a variant of right weakening is proven to hold for both
$\Clarg$ and $\Clwar$.
This implies that (warranted) arguments with a conclusion $x$ account
also as (warranted) arguments for $y$ whenever $\srule{y}{x}$
is present as a non-defeasible rule.
A full analysis of the logical properties of \SDEar\
is outside the scope of this paper; for an in-depth treatment
the reader is referred to~\cite{Cacic2001a}.


\begin{figure}[t]
    \setlength{\unitlength}{.9mm}
\begin{center}
    \begin{picture}(110,55)
    \put(0,5){\framebox(110,50){}}
    \put(45,50){\makebox(0,0){\du{\mathcal{A}}{engine\_ok}}}
    \put(45,46){\makebox(0,0){{\scriptsize (D)}}}
    \put(42,47){\line(-1,-1){10}}
    \put(48,47){\line(1,-1){10}}
    \put(33,34){\makebox(0,0){\du{\mathcal{B}}{\no fuel\_ok}\hspace*{5mm}}}
    \put(33,30){\makebox(0,0){{\scriptsize (D)}}}
    \put(57,34){\makebox(0,0){\hspace*{8mm}\du{\mathcal{E}}{\no engine\_ok}}}
    \put(57,30){\makebox(0,0){{\scriptsize (U)}}}
    \put(30,31){\line(-1,-1){10}}
    \put(36,31){\line(1,-1){10}}
    \put(21,16){\makebox(0,0){\du{\mathcal{C}}{\no low\_speed}\hspace*{4mm}}}
    \put(21,12){\makebox(0,0){{\scriptsize (U)}}}
    \put(45,16){\makebox(0,0){\hspace*{6mm}\du{\mathcal{D}}{fuel\_ok}}}
    \put(45,12){\makebox(0,0){{\scriptsize (U)}}}

    \put(95,50){\makebox(0,0){\du{\mathcal{A}}{engine\_ok}}}
    \put(95,46){\makebox(0,0){{\scriptsize (D)}}}
    \put(95,44){\line(0,-1){10}}
    \put(95,32){\makebox(0,0){\du{\mathcal{E}}{\no engine\_ok}}}
    \put(95,28){\makebox(0,0){{\scriptsize (U)}}}
    \end{picture}
\end{center}
    \caption{Dialectical tree \Tree{\du{\ArgSet{A}}{engine\_ok}}\
             and associated pruned tree \Pruned{\Tree{\du{\ArgSet{A}}{engine\_ok}}}}
    \label{fig:dial_tree_izq}
\end{figure}


\section{\SDEar: theoretical considerations and applications }

\label{sec:applications}

\subsection{Computing Warrant: Bottom-up vs. Top-down}

As described in Section~\ref{sec:defeat_and_war}, the notion of dialectical
tree allows to capture the computation of warranted arguments.
This notion is relevant in the context of defeasible argumentation
in particular, and with respect to scientific reasoning in general.
In most implementations of defeasible argumentation
(e.g. DeLP~\cite{Delp2003}), computation of warrant is performed in
a top-down fashion, based on a depth-first construction of a
dialectical tree. As a marked dialectical tree is
an AND-OR tree, an additional $\alpha$-$\beta$ pruning
can be performed as the tree is built, resulting in a smaller tree,
\emph{pruned} tree.

\begin{example}
Consider  the dialectical label rooted in \du{\ArgSet{A}}{engine\_ok} associated
with the final dialectical label in example~\ref{example:dial}.
This label can be depicted as a \emph{dialectical tree}
as shown in figure~\ref{fig:dial_tree_izq} (left).
The root node of \Tree{\du{\ArgSet{A}}{engine\_ok}}
is labeled as $D$-node.
Note that it is not necessary to compute the whole tree
to mark the root node as $D$.
In fact, considering the pruned tree
\Pruned{\Tree{\du{\ArgSet{A}}{engine\_ok}}}\
shown in figure~\ref{fig:dial_tree_izq} (right),
an equivalent answer would have been obtained.
Note that \Pruned{\Tree{\du{\ArgSet{A}}{engine\_ok}}}\
was obtained from
\Tree{\du{\ArgSet{A}}{engine\_ok}}
by applying $\alpha$-$\beta$ pruning.
\end{example}

The LDS approach provides a bottom-up construction
procedure, as complex labels are built on the basis of
more simple ones.
It can be proven that warrant can be computed by either of these approaches.
In particular, such equivalence result shows that pruning aspects in the top-down
approach (commonly used in implemented argument-based systems as~\cite{Delp2003})
correspond to performing a particular selection of inference rules
in the bottom-up approach.

\begin{theorem}
\label{theorem:equi}
Given an argumentative theory $\Gamma$, the following three cases
are equivalent:
1) The root of \Tree{  \du{\ArgSet{A}}{q} }\ is marked as $U$-node;
2) The root of \Pruned{\Tree{\du{\ArgSet{A}}{q}}},
       is marked as $U$-node; 3)
It is the case that $\Gamma \InferTree \du{\ArgSet{A}}{h}^{U}$.
\end{theorem}

\subsection{Variants of \SDEar}

\begin{figure}[ht]
\begin{center}
\setlength{\unitlength}{0.5mm}
\begin{picture}(160,80)
\put(0,0){
        \begin{picture}(80,80)
            \put(0,-5){\framebox(80,85){}}
            \put(40,76){\makebox(0,0){\SDEsl}}
            \put(40,62){\vector(0,1){8}}

            \put(40,58){\makebox(0,0){\SDEmtdr}}
            \put(40,46){\vector(0,1){8}}

            \put(40,40){\makebox(0,0){\ \SDEskep}}

            \put(28,18){\vector(1,2){8}}
            \put(55,18){\vector(-1,2){8}}

            \put(32,14){\makebox(0,0){\SDEnot}}
            \put(28,3){\vector(0,1){8}}

            \put(32,0){\makebox(0,0){\SDEnlp}}

            \put(60,14){\makebox(0,0){\SDEneg}}
        \end{picture}
          }
\put(80,0){
        \begin{picture}(100,80)
            \put(0,-5){\framebox(80,85){}}
            \put(40,76){\makebox(0,0){\SL}}
            \put(40,62){\vector(0,1){8}}

            \put(40,58){\makebox(0,0){\MTDR}}
            \put(40,46){\vector(0,1){8}}

            \put(40,40){\makebox(0,0){\DELP}}

            \put(30,18){\vector(1,2){8}}
            \put(50,18){\vector(-1,2){8}}

            \put(30,14){\makebox(0,0){\DELPnot}}
            \put(30,3){\vector(0,1){8}}

            \put(30,0){\makebox(0,0){\NLP}}

            \put(60,14){\makebox(0,0){\ \DELPneg}}
        \end{picture}
          }
\end{picture}
\end{center}
\caption{A taxonomy relating the expressive power of \SDEar
         and different argumentation systems}
\label{fig:relacion_poder_expresivo}
\label{fig:one}
\end{figure}
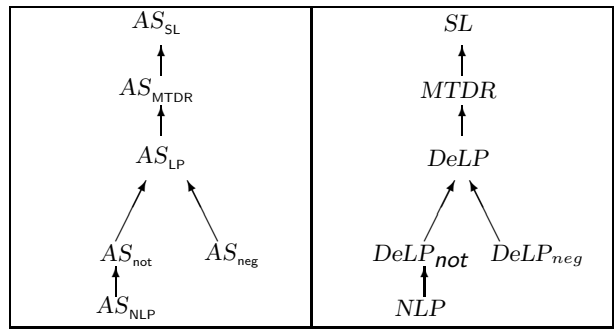

Another interesting issue concerns the definition of \emph{variants}
for \SDEar. Since \SDEar\ is a logical framework, its knowledge-encoding capabilities
are determined by  the underlying logical language, whereas the inference power
is characterized by its deduction rules. Adopting a different knowledge representation
language
or modifying some particular inference rules  would lead to different {variants}
of \SDEar, resulting in a \emph{family} of argumentative systems.
Figure~\ref{fig:one} summarizes some of these variants
of \SDEar\ and their relationship to
to some existing argumentation frameworks, such as
Simari-Loui's~\cite{Simari92},
MTDR (an extension  of the original Simari-Loui approach), 
Defeasible Logic Programming~\cite{Delp2003} and
NLP (normal logic programming), conceptualized in an
argumentative setting as suggested in~\cite{KakasToni99}.
Every variant of \SDEar  is denoted as $AS_{x}$ (standing for
Argumentative System).
Thus, for instance, adopting a restricted first-order language
as the knowledge representation language \LengKR\
leads to \SDEsl , a particular instance of \SDEar\
with a behavior similar to the  argumentative framework proposed in~\cite{Simari92}.
Similarly, restricting the language
\LengKR\ in \SDEar\ to normal clauses~\cite{Lloyd87}
and incorporating an additional inference rule to handle default negation
will result  in a particular argumentative system \SDEnlp, a
formulation similar to normal logic programming (NLP)
under well-founded  semantics
as discussed in~\cite{KakasToni99}.\footnote{A full discussion of different
argumentative frameworks encompassed by \SDEar\ can be found in~\cite{ENGChesnePHD2001}.}
Two distinguished variants of \SDEar  deserved particular attention,
as they allowed to model two particular cases of
defeasible logic programming~\cite{Delp2003},
namely \DELPnot and \DELPneg (\DELP restricted to default and strict
negation, resp.). Such special cases of DeLP could be better understood
and compared in the context of  extensions based on \SDEar.

\subsection{Extending \SDEar\ to incorporate numerical attributes}

The growing success of argumentation-based approaches has
caused a rich crossbreeding  with  other disciplines, providing
interesting results  in different areas such as legal reasoning,
medical diagnosis and  decision support systems.
Many of these approaches rely on \emph{quantitative aspects}
(such as numeric attributes, probabilities or certainty values).
As argumentation provides mostly a non-numerical, \emph{qualitative} setting
for commonsense reasoning, integrating both quantitative and
qualitative features has shown to be highly desirable.

\SDEar\ can be naturally extended to incorporate such quantitative
features, \eg by adding some \emph{certainty factor} $cf$
such that $cf(f)=1$ whenever $f$ corresponds to non-defeasible knowledge,
and $0 < cf(f) < 1$ whenever $f$ stands for defeasible knowledge.
A formula of the form \du{ [\alpha, cf(\alpha) ] }{\alpha}
in the knowledge base $\Gamma$
would therefore stand for ``$\alpha$ is a defeasible formula which
has the certainty factor $cf(\alpha)$''.
Similarly, the formula \du{ [\emptyset, 1 ] }{\alpha}
would stand for ``$\alpha$ is a non-defeasible formula''.
Finally, performing an inference from $\Gamma$
(\ie, building a generalized argument)
would result in inferring a
formula \du{ [\Phi, cf(\Phi) ] }{\alpha},
standing for ``The set $\Phi$ provides an argument for $\alpha$
with a certainty factor $cf(\Phi)$''.

In~\cite{Cacic2002Combining} this approach was  first explored, and
an extension of the  \SDEar\ framework was defined
in order to incorporate numerical attributes.
In this extended framework, deduction rules propagate certainty factors
as inferences are carried out both in arguments and
dialectical trees.\footnote{A detailed analysis of this extension of \SDEar\ is outside the scope of
this paper. For details see~\cite{Cacic2002Combining}.}
It must be remarked that the combination of qualitative and quantitative reasoning
has recently motivated the development of general encompassing
frameworks, such as the one proposed in~\cite{Alcantara2003}, which
allows to deal with default, paraconsistency and uncertainty reasoning,
and is general enough to capture Possibilistic Logic Programs and
Fuzzy Logic Programming, among others.

\section{ Conclusions}
\label{sec:conclu}
\label{sec:conc}

As we have outlined in this paper, Labelled Deductive Systems offer a powerful tool
for formalizing different  aspects of defeasible argumentation.
Many argument-based formalisms exist
(e.g. \cite{Delp2003,PrakkenSartor97,Vreeswijk93}),
relying on a number of shared notions such as
the definition of argument, defeat and warrant.
Such formalisms provided the motivation for the definition of \SDEar,
in which the above notions could be abstracted away by specifying a
suitable underlying logical language and appropriate inference rules.

\SDEar\ provides a formal framework for
argumentative reasoning which can be adapted for different purposes.
As we have detailed in section~\ref{sec:logical},
\SDEar makes it easier to analyze, compare
and relate alternative  argumentative frameworks.
Relevant logical properties of argumentation can also be studied
and analyzed in a formal setting.
Arguments in conflict can be compared and weighed wrt to qualitative
features (\eg specificity) or quantitative ones (\eg certainty factors).
Aggregated preference criteria can be defined to properly
combine these such preference orderings.
The same analysis applies to the construction of dialectical trees.
Alternative approaches can extend the original labeling criterion,
as in the case of considering \emph{accrual of arguments}~\cite{Vreeswijk93,Verheij96}
when assessing  a new certainty factor for the root of a dialectical tree.

In summary, we contend that a general encompassing framework as \SDEar
provides an integrated test-bed for studying different issues
and open problems related to computational models of defeasible argumentation
(such as argumentation protocols,  models of negotiation, resource-bounded reasoning, etc.).
Research in  this direction is currently being  pursued.


\ \\
\noindent
\textbf{Acknowledgments:}
The authors want to thank one of the anonymous reviewers for comments and
suggestions which helped improve the final version of this paper.

\begin{footnotesize}
\bibliographystyle{giia}

\end{footnotesize}

\end{document}